\documentclass[10pt,conference]{IEEEtran}
\IEEEoverridecommandlockouts
\usepackage{cite}
\usepackage{amsmath,amssymb,amsfonts}
\usepackage{algorithmic}
\usepackage{graphicx}
\usepackage{textcomp}
\usepackage{xcolor}
\usepackage{enumitem}
\usepackage{xspace}
\usepackage[acronym]{glossaries}
\usepackage{microtype}
\usepackage{booktabs}
\usepackage{multirow}
\usepackage{array}
\usepackage{adjustbox}
\usepackage{tabularx}
\usepackage{hyperref}
\usepackage{float}
\usepackage{tcolorbox}

\newcommand{\eg}{e.g.,\xspace}
\newcommand{\ie}{i.e.,\xspace}
\newcommand{\etal}{\emph{et al.}\xspace}

\newcommand{\acrostyle}[1]{\textsc{\textls[10]{#1}}}
\newacronym{aia}{\textls[10]{AI\kern0,2em Act}}{Artificial Intelligence Act\xspace}
\newcommand{\AIA}{\gls{aia}\xspace}
\newacronym{hf}{\acrostyle{HF}}{Hugging Face\xspace}
\newcommand{\HF}{\gls{hf}\xspace}
\newacronym{gp}{\acrostyle{GPAI}}{General-Purpose AI models\xspace}
\newcommand{\GP}{\gls{gp}\xspace}

\newtcolorbox{rqanswer}[1]{colback=gray!8,colframe=gray!60,boxrule=0.5pt,arc=1pt,
  left=4pt,right=4pt,top=2pt,bottom=2pt,toptitle=1pt,bottomtitle=1pt,
  coltitle=black,colbacktitle=gray!25,fonttitle=\bfseries,title=#1}

\newif\ifanonymous
\anonymousfalse

\usepackage{twemojis}

\def\BibTeX{{\rm B\kern-.05em{\sc i\kern-.025em b}\kern-.08em
    T\kern-.1667em\lower.7ex\hbox{E}\kern-.125emX}}
\begin{document}

\title{Studying AI Act Mentions: A first look at the impact of the European Union's AI regulation\\ on Hugging Face
}

\ifanonymous
\author{\IEEEauthorblockN{Anonymous Author(s)}}
\else
\author{\IEEEauthorblockN{1\textsuperscript{st} Anamaria Mojica-Hanke}
\IEEEauthorblockA{\textit{University of Passau} \\
Passau, Germany \\
anamaria.mojica-hanke@uni-passau.de}
\and
\IEEEauthorblockN{2\textsuperscript{nd} Tessi Aigbovbiosa-Igbinovia}
\IEEEauthorblockA{\textit{University of Passau} \\
Passau, Germany \\
aigbov01@ads.uni-passau.de}
\and
\IEEEauthorblockN{3\textsuperscript{rd} Steffen Herbold}
\IEEEauthorblockA{\textit{University of Passau} \\
Passau, Germany \\
steffen.herbold@uni-passau.de}
}
\fi

\maketitle

\begin{abstract}
 
The European Union's \AIA  entered into force in August 2024, and its obligations apply in stages. Obligations for providers of general-purpose AI models have applied since August 2025, and transparency rules since August 2026. These obligations affect developers who share and deploy models and applications on public platforms such as \HF, the largest platform for distributing AI artifacts. However, little is known about how practitioners respond to the regulation in practice. To address this gap, we conducted an empirical study of explicit references to the \AIA on \HF. We collected over five million records, including repository cards, discussions, and commits from Models, Datasets, and Spaces, and identified over 1,500 records that explicitly mention the Act. Two coders manually labeled these records using a primarily closed coding approach, and we analyzed the resulting categories per source. We find that explicit references to the \AIA remain rare across all repository types. Repository cards mainly contain claims of compliance and responsibility, like ensuring compliance or assigning responsibility to downstream users, and some claim that the Act does not apply because the model is not a general-purpose AI model. Discussions are used to report alleged violations of the Act, while commits mainly update artifacts that support compliance. Overall, compliance statements focus on who is responsible rather than how compliance is achieved. We believe that this calls for direct platform support, \eg automated checks and documentation templates, as well as clearer guidance on the scope of the \AIA.

\end{abstract}


\section{Introduction}

The European Union's \AIA , Regulation (EU) 2024/1689~\cite{eu_ai_act_2024} (hereinafter also referred to as the {Act}), is the first comprehensive legal framework for artificial intelligence. It entered into force in August 2024, and its obligations apply in stages. The general provisions ({Art.~1--4}) and prohibitions of certain AI practices ({Art.~5}) have applied since February 2025, and obligations for providers of \GP models have applied since August 2025 ({Art.~53}). Since August 2026, the  transparency rules ({Art.~50}) apply, and the European Commission can fine \GP model providers for non-compliance ({Art.~101}). The remaining requirements follow in later years~\cite{eu_ai_omnibus_2026}. The already enforced obligations require providers to document their models and to provide information to downstream users, which directly affects how AI models are shared and reused, among other things.
%

These obligations are particularly relevant for the developers and organizations that share \textit{Models}, \textit{Datasets}, and applications (\textit{Spaces}) on \HF, the largest platform for distributing AI artifacts. The platform hosts over four million public repositories~\cite{huang2026nvidia_hf}: over three million \textit{Models}, over half a million \textit{Datasets}, and over one million \textit{Spaces}. Beyond hosting the files that compose each repository, \HF supports documentation through repository cards, exchange of ideas 
through discussions, and version control through commits.  This makes it a central place where the practical consequences of the \AIA are likely to become visible: owners of the repository may document their compliance in cards,  users may raise concerns in discussions, and contributors may update their repositories in response to the Act.  

Prior work has studied \HF from different angles, from its suitability for empirical studies~\cite{ait2025suitability, jiang2024peatmoss}, maintenance and evolution~\cite{castanio2026}, security and safety~\cite{zhao2024models, yun2026large}, model reuse~\cite{Jiang23, oderinwale2025anatomy}, and governance aspects such as documentation~\cite{liang2024systematic, yang2024navigating}, licensing and copyright~\cite{longpre2024large, stalnaker2026empirical}. The few studies that relate \HF to the \AIA analyze a specific subset of data and specific provisions, such as dataset cards against data-governance requirements~\cite{kapitsaki2026regulatorycompliance} or model cards against technical-documentation requirements~\cite{mamirov2025aitransparency}. However, little is known about how the community is responding to the \AIA in practice.  Understanding this matters for practitioners, who must interpret and act accordingly to the new obligations, for platform designers, who could better support compliance; and for policymakers, who rely on such evidence to refine guidance, codes of practice, and implementing measures.

To address this gap, we conduct an empirical study of explicit references to the \AIA on \HF. We collected all public \textit{Models}, \textit{Datasets}, and \textit{Spaces} from \HF, together with their repository cards, discussions, and commits, and identified over 1,500 records mentioning the Act. Two coders manually classified each record into six categories: \textit{compliance and responsibility statements}, \textit{non-applicability claims}, \textit{compliance artifacts}, \textit{knowledge resources}, \textit{reported violations} and \textit{false positive}. We then analyzed the distribution of these categories and illustrated each with examples. Our study is guided by three research questions:

\begin{enumerate}[label=\textbf{RQ\arabic*:}, leftmargin=*, align=left]
    \item How do repository cards reference the AI Act? 
    \item What AI Act-related concerns do users raise in discussions? 
    \item What AI Act-related changes do commits describe? 
\end{enumerate}

\noindent
The contributions of our study are:
\begin{itemize}
    \item The first study to systematically examine explicit references to the \AIA across repository cards, discussions, and commits, covering more than 4.2 million cards, one million discussions, and 73 million commits.
    \item We found that explicit references are rare and mostly statements about who is responsible for compliance rather than how compliance is achieved. Models mainly document their own artifacts, while spaces build tools and resources around the regulation. 
    \item We present implications for developers and platforms, arguing that platform-level support, such as automated checks and documentation templates, and clearer guidance on the scope of the \AIA are needed.
\end{itemize}

The rest of the paper is organized as follows. Section~\ref{sec:background} presents background and related work, and Section~\ref{sec:method} describes our methodology in detail. Section~\ref{sec:results} reports the results, which we discuss in Section~\ref{sec:disc}, together with their implications in Section~\ref{sec:implications}. Section~\ref{sec:limitations} addresses threats to validity, and Section~\ref{sec:conclusions} concludes the paper and outlines directions for future work.

\section{Background and Related Work}
\label{sec:background}
In this section, we briefly describe the \AIA and the most relevant articles for this study, as well as previous work on \HF that analyzes aspects that the \AIA nowadays regulates, \eg documentation. 

\subsection{The AI Act}
The \AIA~\cite{eu_ai_act_2024} regulates AI by risk level. Several of its provisions can affect developers who share models on platforms such as \HF. Examples include the prohibition of certain AI practices (Art.~5), which regulates specific uses rather than \GP. Requirements for high-risk systems, such as data governance and technical documentation (Arts.~10--11), which were recently postponed to 2027--2028~\cite{eu_ai_omnibus_2026}, or additional duties for general-purpose models with systemic risk (Art.~55), which only concern a small number of the most capable models. We focus on the two provisions that apply most directly to the majority of models and applications shared on the platform: Art.~53 on \GP and Art.~50 on transparency. \textit{For the purposes of this study, the following headings refer to the practical purpose of each provision rather than to the official article titles in the Act.}


\textbf{Art.~53: documenting general-purpose models \GP.} Since August 2025, developers who release a \GP must document it following Annexes XI and XII, which regulate the technical documentation and the documentation for downstream providers, respectively. In practice, this means that developers must present technical documentation of the model; usage information for developers who build on it; a copyright policy, including respecting opt-outs from data-mining; and a public summary of the training data, based on a template from the EU AI office~\cite{ec2025_summary_template}. Note that free open-source models\footnote{Open-source model: models that allow the access, usage, modification, distribution and that have publicly available the parameters, weights, model architecture, and information on model usage.} without systemic risks\footnote{Under Art.~51, ``systemic risk'' is automatically triggered if a model's total training compute exceeds $10^{25}$ FLOPs or if the EU AI Office manually designates it.} get a partial exemption: only the copyright policy and training-data summary are required.

\textbf{Art.~50: labeling AI interactions and generated content.} From August 2026, it should be clear for the users when they are interacting with AI systems, for example when they are interacting with a chatbot.  Systems that generate images, audio, video, or text must mark their outputs in a machine-readable way, for example with a watermark or metadata. Deep fakes must also be disclosed as such. For \HF, this concerns both \textit{Models} and \textit{Spaces} (demo applications) built on them.

The two provisions converge on 2~August 2026. From that date, the transparency obligations of Art.~50\footnote{Art.~111: Generative systems already on the market before 2~August 2026 have until 2~December 2026 to implement the machine-readable marking required by Art.~50~(2)~\cite{eu_ai_act_2024,eu_ai_omnibus_2026}.} apply, and the Commission can fine \GP providers who do not comply with Art.~53. It is therefore from this point that we can reasonably expect developers on \HF to react to the regulation (in particular for Art.~53): discussing the new obligations and adapting their repositories. Section~\ref{sec:method} describes how we use this date to define our data collection.

\subsection{Related Work}

Current related work on \HF covers different aspects, such as its suitability/dataset collection for empirical studies~\cite{ait2025suitability, jiang2024peatmoss}, the characterization of model reuse~\cite{Jiang23, oderinwale2025anatomy}, security-related aspects like identifying malicious models~\cite{zhao2024models} or key leakage~\cite{yun2026large}, how models change over time~\cite{castanio2026}, and documentation and governance~\cite{liang2024systematic, yang2024navigating}.

Only a few recent studies assess \HF artifacts directly against the \AIA. Kapitsaki~\cite{kapitsaki2026regulatorycompliance} analyzed 11,682 popular \textit{Dataset} cards against data-governance requirements related to Art.~10, finding that only 3.47\% describe their source data and 6.81\% include a bias and limitations section. Mamirov~\etal~\cite{mamirov2025aitransparency} scored 100 \textit{Model} cards against the Annex~IV documentation requirements, and found that technical details were usually present, while safety and interpretability information was largely missing. 

Earlier studies, conducted before these obligations applied, examined documentation and licensing practices that the \AIA now regulates.  Liang~\etal~\cite{liang2024systematic} analyzed 32,111 model cards and found that fewer than half of the models have a  card, and sections such as limitations and environmental impact are rarely filled (\ie related to Annex~XI of the \AIA). Yang~\etal~\cite{yang2024navigating} analyzed 7,433 dataset cards and found that 86\% of the top 100 most downloaded datasets have documentation, compared with only 7.9\% of the cards of datasets with no downloads. Licensing, which determines both the copyright obligations of Art.~53~(1)(c) and eligibility for the open-source exemption of Art.~53~(2), is similarly incomplete. Longpre~\etal~\cite{longpre2024large} found license omission rates above 70\% for the datasets they audited on \HF and GitHub, and Stalnaker~\etal~\cite{stalnaker2026empirical} found that only 37.8\% of models and 27.6\% of datasets on \HF declare a machine-readable license.

Overall, prior work shows that much of the information the \AIA requires is missing from or incomplete in \HF artifacts. However, most of these studies were conducted before the \AIA applied, and the few that explicitly reference it assess only specific provisions for a single type of artifact (\eg model or dataset cards). Moreover, to the best of our knowledge, no study has examined how the \HF community discusses, interprets, or acts on the regulation with a broader perspective.

\section{Methodology}
\label{sec:method}

\begin{figure}
\centerline{\includegraphics[width=\columnwidth]{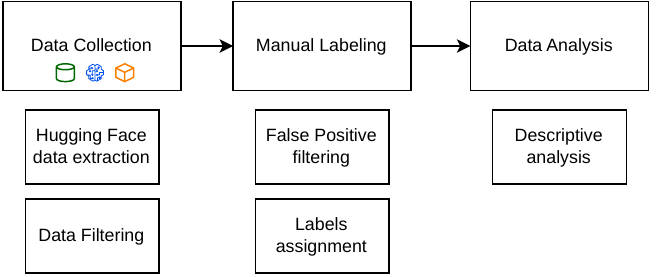}}
\caption{General Method}
\label{process}
\end{figure}

This study follows three main steps (see Fig.~\ref{process}): (A) we collect content from Hugging Face (HF) from the three main repository types (\ie models, datasets, and spaces) and filter the content that explicitly mentions the \AIA; (B) we manually classify each match to understand how the \AIA is being mentioned; and (C) we analyze the resulting labels per source (repository cards, discussions, and commits) to answer RQ1--RQ3.

\subsection{Data Collection}

We collected data from the three main types of repositories\footnote{We excluded \textit{buckets} since they are non-versioned storage without commits or repository cards.} on \HF: \textit{Models}, \textit{Datasets}, and \textit{Spaces}.  Each of the selected types relates to the \AIA in a different way:

\begin{itemize}
    \item \textbf{\textit{Models}:} The \AIA imposes obligations on models, in particular on providers of general-purpose AI models, including technical documentation (Art.~53). Model repositories are where such models are published, documented, and discussed.
    \item \textbf{\textit{Datasets}:} Datasets are commonly used to train models. The \AIA addresses training data, for instance through the requirement to publish a summary of training content and to comply with copyright law (Art.~53~(1)).
    \item \textbf{\textit{Spaces}:} Spaces  host applications and demos built on top of models and datasets. Since they typically take user input and generate outputs such as text, images, or predictions, some of them may qualify as AI systems in the sense of the \AIA (Art.~3~(1)~\cite{eu_ai_act_2024}), for which further obligations, such as transparency obligations (Art.~50), may apply depending on their purpose and use.  
\end{itemize}
From each repository type, we collected three data sources: repository cards (\ie \texttt{Readme.md} file), discussions and commit metadata.

\begin{itemize}
    \item \textbf{Repository cards:}  On \HF, the repository card is the main means of documenting a repository and therefore the most likely place to state compliance with, or the applicability of, the \AIA.
    \item \textbf{Discussions:} On \HF, discussions (including pull requests) play a similar role to issues on collaborative platforms such as GitHub. We therefore analyzed them to identify what users report or debate about the \AIA.  We treated each thread, consisting of the initial post (parent) and its comments (events), as a single unit of analysis. That is, if either the initial post or any of its comments matched our search patterns, we analyzed the whole thread. Note that events also include changes in the metadata of discussions, such as title changes, and commits associated with pull requests~\cite{hf_hub_pr_docs} as they are built on Git references~\cite{ait2025suitability}.
    \item \textbf{Commits:} Since our goal is to understand what is being discussed/referenced about the \AIA, rather than the code changes the regulation may have triggered, we collected only commit titles and messages. 
\end{itemize}


\textbf{\HF data extraction.} We retrieved the data using the \HF Hub API~\cite{hf_hub_api}, inspired by the HFCommunity tool~\cite{ait2024hfcommunity}, collecting all public repositories created between the launch of the Hub (January 2016~\cite{mehta2026nvidia})\footnote{We use the launch of the Hub as the starting point because the earliest repositories (before 02.03.2022~\cite{hfdocumentation}) do not have a creation date in the API metadata.} and 2~September 2026. We chose this end date because 2~August 2026 is when the transparency obligations of Art.~50 started to apply and when \GP providers became subject to fines for non-compliance with Art.~53 (one year after the enforcement date, see Section~\ref{sec:background}). Collecting data one month later gives practitioners time to discuss the new obligations and to adapt their \textit{Models}, \textit{Datasets}, and \textit{Spaces} accordingly. Rather than sampling, we collected and stored public repository data available from the Hub and then selected those matching our criteria. Since the extraction started after 2~September 2026 and took approximately three weeks, we removed all repositories, commits, and discussion events created after that date. Repository cards, however, reflect their state at extraction time and may therefore include changes made up to 27~September 2026.

\begin{table}[htbp]
  \centering
  \small
  \caption{Counts by source and repository type.}
  \label{tab:source-repo}
  \begin{adjustbox}{max width=\columnwidth}
  \begin{tabular}{l r r r}
    \toprule
    & \multicolumn{3}{c}{\textbf{Repository type}} \\
    \cmidrule(lr){2-4}
    \textbf{Source}  & \textbf{Models} & \textbf{Datasets} & \textbf{Spaces} \\
    No. Unique               & {3,030,698} & {1,028,533} & {1,461,709} \\
    \midrule
    \textbf{Card}                 & 2,012,244 (66.4\%) & 731,062 (71.1\%) & 1,454,679  (99.5\%) \\
    \multicolumn{4}{l}{\textbf{Discussions}} \\
    \quad Parent        & 425,345 & 520,967 & 119,432 \\ 
    \qquad { unique}    & 240,394 (7.9\%)&  349,791 (34.0\%)& 30,933 (2.1\%)\\ 
    \quad Comments      & 1,201,226 & 1,218,473 & 325,322   \\ %
    \qquad { unique}    &  240,252 (7.9\%)& 347,973 (33.8\%) & 30,923 (2.1\%) \\ %
    \addlinespace
    \textbf{Commits}    &  32,402,190&  25,081,333&  14,192,775\\ 
    \quad { unique}     &  2,964,309 (97.8\%)&  981,152 (95.4\%)&  1,355,714 (92.7\%)\\ 
    
    \bottomrule
  \end{tabular}
   \end{adjustbox}

  \vspace{2pt}
    \parbox{\columnwidth}{\footnotesize \textbf{Note:} For each source, the first row gives the total number of items, and the \emph{unique} row gives the number and percentage of repositories with at least one.}
\end{table}

Table~\ref{tab:source-repo} summarizes the number of records per repository type and source. As expected, almost all \textit{Spaces} have a repository card (99.5\%), since a card with YAML configuration is required to run them~\cite{YAMLHF}. \textit{Models} and \textit{Datasets} have a higher share of repository cards than reported in previous work~\cite{liang2024systematic}. Note, however, that we only check whether a  \texttt{Readme.md} exists, not whether it has meaningful content. 
Finally, a small share of repositories (2.2--7.3\%) have no commits in our dataset, because they were deleted, gated (\textit{Models} and \textit{Datasets} only), or made private between the collection of their IDs and the retrieval of their commits. The same applies to cards and discussions, so these counts should be read as lower bounds.



\textbf{Data Filtering.} To identify relevant records, we created full-text indexes in MariaDB on the relevant text fields of each data source (\ie repository cards: card content; discussions: title, summary, and content; commits: title and message)  and queried them for the patterns listed in Table~\ref{tab:term-matching2}. Matching was case-insensitive, and the full-text parser treats punctuation as a word separator, so the variants shown in the table, as well as forms such as \textit{AI-Act}, \textit{AI.Act}, or \textit{Regulation (EU) 2024/1689}, were matched by the same patterns. Since MariaDB does not index words with fewer than four characters by default, we set the minimum token size to two so that short terms such as \textit{AI} and \textit{Act} were indexed.
\begin{table}[htbp]
  \centering
  \small
  \caption{Search patterns and examples of matching strings.}
  \label{tab:term-matching2}
  \begin{tabularx}{\columnwidth}{l >{\raggedright\arraybackslash}X}
    \toprule
    \textbf{Pattern} & \textbf{Examples of matching strings} \\
    \midrule
    AI ACT & ai act, eu ai act, european ai act \\
    \addlinespace
    AIA & aia \\
    \addlinespace
    artificial intelligence act & artificial intelligence act, eu artificial 
      intelligence act, european artificial intelligence act \\
    \addlinespace
    2024 1689 & 2024 1689, regulation eu 2024 1689, regulation 2024 1689 eu
    \\
    \bottomrule
  \end{tabularx}
    \par\vspace{3pt}
    \parbox{\columnwidth}{\footnotesize \textbf{Note:} The examples are illustrative and not exhaustive; each pattern also matches capitalization and punctuation variants.}
\end{table}

\subsection{Manual Classification}

For the manual labeling, we followed a primarily closed coding methodology~\cite{seaman1999qualitative, saldana2009coding}. Two coders independently labeled all the 2,000 cards, 73 discussions, and 1,259 commits, assigning each data point one label. The categories were defined before labeling based on the ways we expected the \AIA to be addressed on \HF: statements about compliance, resources for understanding it, tools for assessing it, and complaints, except for \textit{non-applicability claims}, which emerged during the labeling process. When this category was introduced, all previously labeled records were relabeled by both coders.

\begin{itemize}
    \item \textbf{Compliance and responsibility statements.} The record makes a claim about the repository’s relationship to the \AIA: that it complies, or that responsibility for legal aspects (\eg copyright, data, use, transparency) lies with the owner, a third party, or the user. In discussions, this covers debates about whether a repository is or will be subject to the Act. In commits, it covers changes to the \AIA-related parts of the card, as indicated in the title or message. 

    \item \textbf{Non-applicability claims.} The record states that the \AIA does not apply to the repository, typically because the model is not a \GP, or it has not been certified. 
    
    \item \textbf{Knowledge resources.} The \textit{Model}, \textit{Dataset}, or \textit{Space} serves to understand the \AIA, for example a Q\&A model trained on the Act, a dataset containing the Act among other regulations, or a benchmark evaluating popular models against the Act. 
    
    \item \textbf{Compliance artifacts.} The repository provides a means to assess \AIA compliance, such as logging mechanisms, compliance checkers, or datasets for checking whether a model complies. 
    In commits, this covers changes to such artifacts, as indicated in the title or message. 

    \item \textbf{Reported alleged violations.} Someone reports or identifies a potential violation of the \AIA. These are claims by users, not violations verified by the authors.
    
    \item \textbf{False positive}. The match does not refer to the \AIA (\eg ``AIA'' used as an acronym for something else
    ), the record was not in English, or the \AIA is only mentioned incidentally, for instance as an example in a repository whose main purpose is unrelated to the Act or compliance. If supporting  \AIA compliance or understanding the regulation is the main purpose of the repository, the record is instead labeled as a \textit{compliance artifact} or \textit{knowledge resource}, respectively. 
\end{itemize}

During the labeling, the coders also excluded records not written in English  (language filtering). After both coders had finished, we computed inter-rater agreement using Cohen’s kappa~($\kappa$)~\cite{cohen1960coefficient} on the independently assigned labels. The coders then met to discuss all disagreements and agreed on a single label for each record. 

We obtained an overall $\kappa$ of 0.93 (0.91, 0.79, and 0.89 for cards, discussions, and commits, respectively) and an overall percent of agreement of 96.01\% (95.15\%, 84.93\%, and 98.01\% for cards, discussions, and commits, respectively). In total, we excluded 1,764 records across all data sources and repository types (\ie the records were either false positives and/or non-English records), resulting in the final dataset summarized in Table~\ref{tab:term-matching}. Most false positives came from the ``AIA'' pattern  (1,400). They were mainly caused by repositories working with the Arosi language, whose ISO 639-3 code is \texttt{aia}, and by cards containing incorrectly encoded text (\ie mojibake), where character sequences such as ``\~A\~\i a'' were matched as ``AIA'' because our search was case- and accent-insensitive.

\begin{table*}[t]
  \centering
  \footnotesize
  \caption{Number of record matches per search pattern, source and repository type.}
  \label{tab:term-matching}
  \setlength{\tabcolsep}{5pt}
  \begin{tabular}{l rrr rrr rrr}
    \toprule
    & \multicolumn{3}{c}{\textbf{Cards}}
    & \multicolumn{3}{c}{\textbf{Discussions}}
    & \multicolumn{3}{c}{\textbf{Commits}} \\
    \cmidrule(lr){2-4} \cmidrule(lr){5-7} \cmidrule(lr){8-10}
    \textbf{Pattern (example matches)} & M & D & S & M & D & S & M & D & S \\
    \midrule
    AI Act \textit{(EU AI Act, European AI Act)}
      & 1,077 & 375 & 250 & 30 & 17 & 8 & 41 & 12 & 113 \\
    AIA
      & 115 & 96 & 84 & 5 & 12 & 0 & 40 & 1,011 &  42\\
    Artificial Intelligence Act \textit{(EU Artificial Intelligence Act)}
      & 6 & 8 & 4 & 1 & 0 & 0 & 0 & 0 & 0 \\
    2024 1689 \textit{(Regulation (EU) 2024/1689)}
      & 470 & 16 & 14 & 3 & 1 & 1 & 0 & 0 & 4 \\
    \midrule
    \textbf{Unique records} & 1,195 & 470 & 335 & 36 & 29 & 8 & 81 & 1,023 & 155 \\
    \textbf{Total (M+D+S)} & \multicolumn{3}{r}{2,000} & \multicolumn{3}{r}{73} & \multicolumn{3}{r}{1,259} \\
    \cmidrule{1-10}
    \textbf{Unique true positive (tp)} & 1,012 & 211 & 171 & 29 & 7 & 7 & 40 & 6 & 86 \\
    \textbf{Total unique tp (M+D+S)} & \multicolumn{3}{r}{1,394} & \multicolumn{3}{r}{41} & \multicolumn{3}{r}{132} \\
    
    \bottomrule
    \multicolumn{10}{c}{\rule{0pt}{10pt}\textbf{M:} \textit{Models}, \textbf{D:} \textit{Datasets}, and \textbf{S:} \textit{Spaces}} \\
  \end{tabular}
\end{table*}

\subsection{Data Analysis}

We analyzed each source separately, in line with our three research questions: repository cards for \textit{$RQ_1$}, discussions for \textit{$RQ_2$}, and commits for \textit{$RQ_3$}. For each source, we report the distribution of categories in general and per repository type, \ie \textit{Models}, \textit{Datasets}, and \textit{Spaces}, and
illustrate each category with examples.

\section{Results}
\label{sec:results}

Explicit mentions of the \AIA are rare. Out of more than four million repository cards, one million discussions, and 73 million commits in our dataset, only a small fraction mention the regulation. After removing false positives and non-English items, we retained 1,394 repository cards, 41 discussions, and 132 commits as true positives (Table~\ref{tab:term-matching}, \emph{Total unique tp}). In the following subsections, we report the labeling results for repository cards ($RQ_1$), discussions ($RQ_2$), and commits ($RQ_3$) ($n=$ No. of records; respective percentage\%).\footnote{Percentages may not sum to 100\% due to rounding.}

\subsection{\texorpdfstring{$RQ_1$}: How do repository cards reference the AI Act?}
\begin{figure}[H]
\centerline{\includegraphics[width=\columnwidth]{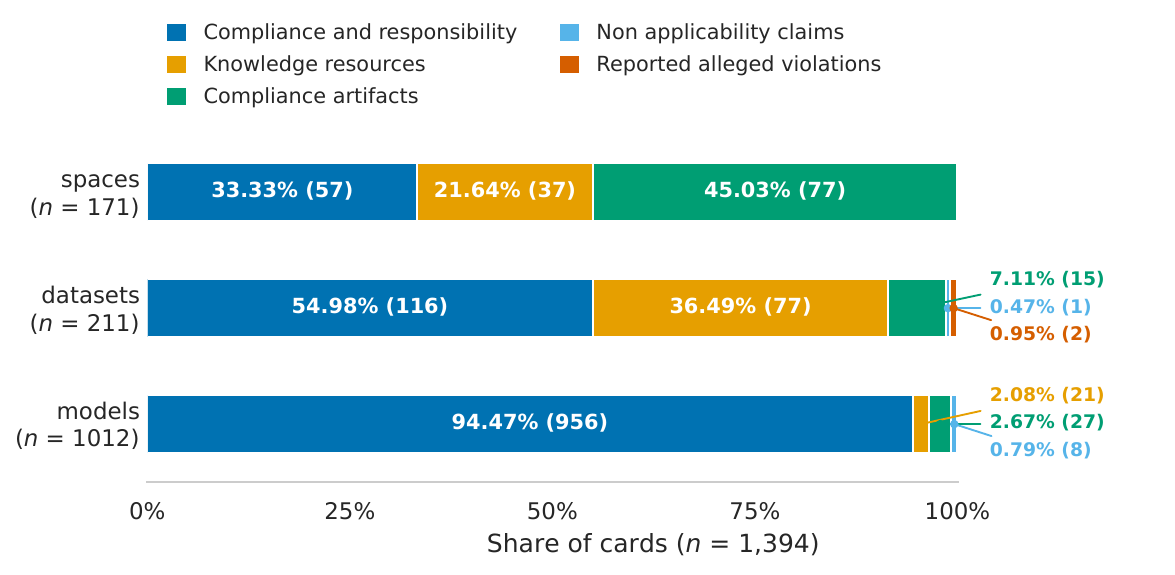}}
\caption{Unique cards that explicitly mention the AI Act}
\label{fig:rq1}
\end{figure}

Figure~\ref{fig:rq1} depicts the distribution of labels across repository types for the 1,394 repository cards that explicitly mention the \AIA. Most of these cards belong to \textit{Models} (1,012; 72.54\%), followed by \textit{Datasets} (211; 15.12\%), and \textit{Spaces} (171; 12.25\%). 

\vspace{0.5cm}

Across all cards, \textit{Compliance and responsibility statements} is by far the most frequent label (1,129), followed by \textit{Knowledge resources} (137). \textit{Non-applicability claims} (9) and \textit{Reported alleged violations} (3) are rare. Repository cards are the only source in which all five labels occur, and the only one containing \textit{Non-applicability claims}.

The distribution of labels differs considerably across repository types. In \textit{Models}, \textit{Compliance and responsibility} accounts for 94.36\% of the \AIA mentions (955), while the remaining three labels together account for 5.64\% (57). \textit{Datasets} show a more mixed picture, \textit{Compliance and responsibility} (116; 55.97\%) is again the most frequent label, followed by \textit{Knowledge resources} (77; 36.49\%), with the remaining 8.5\% spread across \emph{Compliance artifacts}, \textit{Non-applicability claims}, and \textit{Reported alleged violations}. \textit{Spaces} are the most balanced, \textit{Compliance artifacts} is the most frequent label (75; 43.85\%), followed by \textit{Compliance and responsibility} (57; 33.33\%) and \textit{Knowledge resources} (39; 22.80\%). Spaces contain neither \textit{Non-applicability claims} nor \textit{Reported alleged violations}, whereas both labels appear in \textit{Models} and \textit{Datasets}.

To illustrate the labels, we present representative examples from the labeled repository cards. These examples are illustrative and do not constitute a systematic sub-classification of each label. 

\textit{Compliance and responsibility statements} typically consist of a single sentence/paragraph. For example, owners affirm that their own artifact complies with the \AIA: ``The licence aims to support open and responsible downstream use while ensuring compliance with applicable requirements related to copyright, personal data protection, and the EU \textit{AI Act}''. 
In other cases,  owners shift the responsibility for compliance to the downstream users/upstream developers of the artifact: ``Users in the EU acknowledge that their use of this model is subject to applicable EU regulations, including the AI Act and DSA provisions''. 
Some owners also classify  their artifact according to the provisions of the \AIA: ``This system may qualify as a high-risk AI system under EU \textit{AI Act} Annex III''. 
In other cases, the mention of the \AIA introduces a section with further information supporting compliance, such as ``This adapter is provided as a fine-tuned LoRA under the \textit{AI Act} framework (Regulation EU \textit{2024/1689})''. Across these statements, when the \AIA is mentioned, specific provisions are also referenced, such as Art.~50, Art.~53, Art.~14, Art.~5, among others. 

In contrast, \textit{Non-applicability claims} state that the \AIA, or parts of it, does not apply to the artifact. Owners stated, for example, that their repository is not certified under the \AIA: ``UNCERTIFIED is the default: no competent authority exists to confer EU \textit{AI Act} conformity, so neither can we.'' Others argued, based on the definitions in the Act, that its obligations do not apply to their artifact. ``On that basis, the release is preliminarily assessed as not falling within the provider obligations for GPAI models under the EU \textit{AI Act} definitions''. One owner argued that their model is not a \GP model because its training compute falls below a specific threshold attributed to the Act, although the \AIA itself defines \GP models without a compute threshold (Art.~3~\cite{eu_ai_act_2024}); such a threshold only appears in a legally non-binding guideline by the European Commission~\cite{gpai-guidelines}: ``This model is not a ``general-purpose AI model'' (GPAI) under the EU \textit{AI Act}. The Act's GPAI classification threshold targets models trained with more than 10²³ FLOPs [\ldots] ''.


\textit{Knowledge resources} use the \AIA as content that the artifact contains or builds on it (\eg uses it as training data or parts of it).  For example, one model card states that the model ``utilizes the European AI Act regulation text as training data in three languages''. In other cases, the \AIA appears alongside further regulatory frameworks: ``Documents covered	EU AI Act, NIST AI RMF GenAI Profile, NIS2, [\ldots]''. The  regulation can serve as a knowledge base for answering questions, such as a ``Hybrid Knowledge-Graph + RAG engine that answers compliance questions across GDPR, the EU AI Act, [\ldots]''.

\textit{Compliance artifacts} are artifacts intended to support compliance with the \AIA. For example, some artifacts detect or audit specific aspects regulated by the \AIA, such as personal data: ``A multilingual [personally identifiable information] detector built for EU compliance reality: [\ldots], AI Act high-risk data''. Others audit the documentation of models: ``[\ldots] AI agent acts as a responsible AI auditor, reviewing ML model cards for completeness, [\ldots] and regulatory compliance against frameworks like the EU \textit{AI Act} and NIST AI RMF''. This label also includes datasets for evaluating agents in compliance tasks, such as ``[\ldots] 80K multi-turn tool-calling traces for EU regulatory compliance workflows — GDPR, \textit{AI Act}, [\ldots]''.


Finally, \textit{Reported alleged violations} are cards in which the owners point out potential non-compliance of other/own artifacts with the \AIA. This label is rare (two cards). For example, a dataset card that documents
the licensing status of several corpora notes that one of them has ``no license tag on the Hub'' and argues that ``[u]nder EU \textit{AI Act} GPAI obligations and most enterprise procurement policies, an undocumented corpus is unusable without an audit''.

\vspace{-5pt}
\begin{rqanswer}{Answer to RQ1}
\textit{Model} cards reference the \AIA mainly through short \textit{Compliance and responsibility} statements, in which owners affirm compliance, shift responsibility to downstream users/upstream developers, or classify their artifact under the regulation. In \textit{Spaces}, the \AIA is more often the subject of the artifact itself, serving as the basis for \textit{Compliance artifacts} and \textit{Knowledge resources}, while \textit{Dataset} cards show a mix of both. 
\end{rqanswer}


\subsection{\texorpdfstring{$RQ_2$}: What AI Act-related concerns do users raise in discussions?}
\begin{figure}[H]
\centerline{\includegraphics[width=\columnwidth]{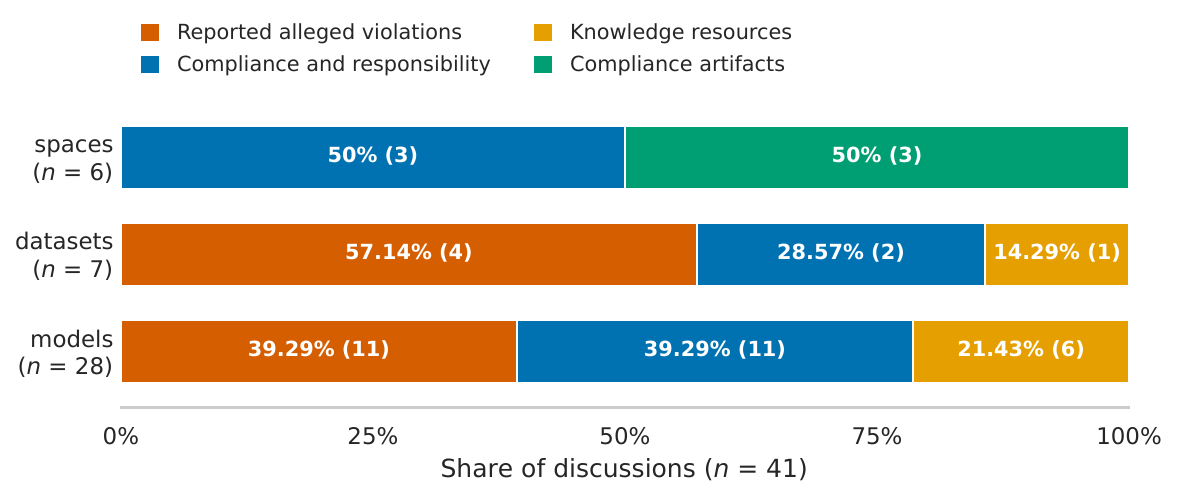}}
\caption{Unique discussions that explicitly mention the AI Act}
\label{fig:rq2}
\end{figure}

Figure~\ref{fig:rq2} shows the distribution of labels across repository types for the 41 discussions that explicitly mention the \AIA. Most of these discussions take place in \textit{Models} (28; 68.29\%), while \textit{Datasets} (7; 17.07\%) and \textit{Spaces} account for six (14.63\%).

\vspace{0.5cm}
\vspace{0.5cm}

As in repository cards, \textit{Compliance and responsibility statements} is the most frequent label in discussions (16; 39.02\%). These discussions take different forms. Some users point to upcoming obligations as a reason for future developments: ``The EU AI Act enforcement begins August 2026. Art. 9 mandates risk management systems for high-risk AI [\ldots]''. Others respond to compliance statements made by the owners: ``Your statement that `Apertus was developed with due consideration to Swiss data protection laws, [\ldots] and the transparency obligations under the EU AI Act' is appreciated''. Owners also use discussions to justify previous changes motivated by the \AIA, ``We have officially deprecated Stable Diffusion 2.0 and 2.1. This is part of our effort to clean up and consolidate our model offering and to get ahead of upcoming compliance requirements for the EU AI Act in 2026''.

In contrast to repository cards, where they are rare, \textit{Reported alleged violations} are the second most frequent label in discussions (15; 36.58\%). Ten of these discussions were opened by the same user, each pointing out a possible risk, for example that operating ``without strict data governance protocols exposes the infrastructure to severe liability under the upcoming EU AI Act baselines (Article 10)''. In two of these cases, the repository owners hid the discussion, leaving only its title visible, and in three cases, subsequent comments reported the user as spam. Four of the remaining
discussions concern licensing and copyright issues, for example: ``the Hub repo declares no license tag, and there is no LICENSE file in the repo. For researchers that's fine, but under EU AI Act GPAI documentation duties [\ldots]'' or ``[\ldots] you are actively facilitating non-compliance with the EU AI Act''. The last discussion concerns the ethics of a model and refers to the \AIA as an authority that enforces ethical requirements.


\textit{Knowledge resources} appear in seven discussions (17.07\%). In these discussions, users share or propose material that explains the \AIA. For example, in two discussions of the same artifact, users presented a blog post explaining the regulation and its implications: ``\textit{AI Act}: Bringing AI Systems into Compliance [\ldots]''.  In other cases, users point to models they have built to help understand the \AIA, ``Hey I have build the EU \textit{AI act} model on the top of google/gemma2b [\ldots]''. Others propose changes to one of the main repositories of Hugging Face, which users rely on as a resource for understanding the \AIA: ``Add OS Developers Guide to the EU \textit{AI Act} blogpost''.

\textit{Compliance artifacts} are rare, occurring only three times and only in \textit{Spaces} (7.31\%). These discussions present tools intended to support compliance with the \AIA, for example a tool for adding transparency metadata to AI-generated content, which its developers link to the obligations under Art.~50 (discussion on a space), or an agent described as an ``EU AI Act Compliance Agent [\ldots]''.

As for repository cards, the distribution of labels differs across repository types. In \textit{Models}, \textit{Reported alleged violations} and \textit{Compliance and responsibility statements} occur equally often (11 each; 39.29\%), followed by \textit{Knowledge resources} (6; 21.43\%). In \textit{Datasets}, four of the seven discussions report alleged violations (57.14\%), two contain \textit{Compliance and responsibility statements} (28.57\%), and one refers to a \textit{Knowledge resource} (14.3\%). \textit{Spaces} is the only repository type without \textit{Reported alleged violations} or \textit{Knowledge resources}: half of their discussions contain \textit{Compliance and responsibility statements} and the other half \textit{Compliance artifacts} (3 each), the latter label appearing only in this repository type.

\begin{rqanswer}{Answer to RQ2}
Discussions mentioning the \AIA are rare and mostly take place in \textit{Model} repositories. Users raise two main kinds of concerns at a similar frequency: \textit{Compliance and responsibility statements}, in which users point to upcoming obligations or respond to owners' compliance claims and owners explain changes motivated by the \AIA, and \textit{Reported alleged violations}, mostly concerning missing licenses and copyright. Unlike in repository cards, alleged violations are frequent, although most of them stem from a single user. 
\end{rqanswer}

\subsection{\texorpdfstring{$RQ_3$}: What AI Act-related changes do commits describe?}

\begin{figure}[H]
\centerline{\includegraphics[width=\columnwidth]{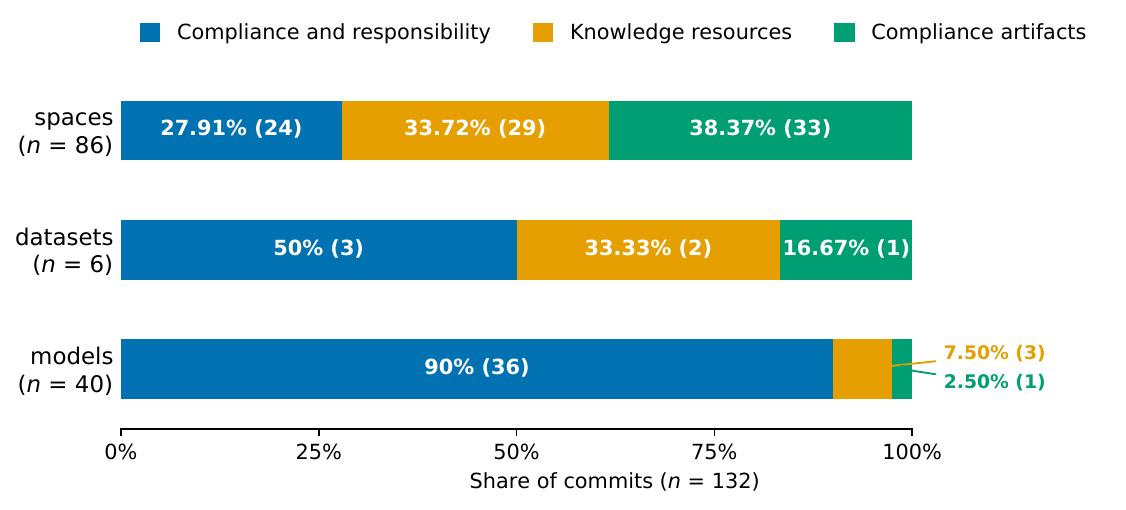}}
\caption{Unique commits that explicitly mention the AI Act}
\label{fig:rq3}
\end{figure}

Figure~\ref{fig:rq3} displays the distribution of labels across repository types for the 132 commits that explicitly mention the \AIA. Unlike repository cards and discussions, most of these commits belong to \textit{Spaces} (86; 65.15\%), followed by \textit{Models} (40; 30.30\%) and \textit{Datasets} (6; 4.54\%).

As in repository cards and discussions, \textit{Compliance and responsibility statements} is the most frequent label in commits (63; 47.72\%). Some simply state compliance with the \AIA, such as ``[\ldots] GDPR + EU \textit{AI Act} compliance [\ldots]''. Others describe the modification of compliance statements, for example ``docs: append EU \textit{AI Act} §50 transparency section [\ldots]'', or metadata such as an ``[\ldots] disclosure-safe aggregate outcomes for Russia policy framing, EU AI Act identifier [\ldots]''. In other cases, commits describe changes made to meet requirements of the \AIA, such as ``satisfying EU \textit{AI Act} audit trail requirements''.


\textit{Compliance artifacts} appear in 35 commits (26.51\%), almost all of them in \textit{Spaces} (33). Some commits describe the intended use and the limits of the artifact, for example a tool ``[\ldots] [i]ntended as a first-pass triage tool to support (not replace) human compliance review under the EU AI Act'', or add ``[\ldots] a disclaimer section regarding limitations of the EU AI Act compliance tool''. Others document the functionalities of the artifact, ``This section proves \textit{AI Act} Navigator delivers what clients need: Professional, actionable compliance roadmaps [\ldots]'', or record actions taken with it, such as ``Deploy AI Act Compliance app''.

\textit{Knowledge resources} appear in 34 commits (25.75\%), most of them in \textit{Spaces} (29). Some commits describe assistants for querying regulatory texts, for example, ``[\ldots] It helps users query European regulations ([\ldots] EU \textit{AI Act} [\ldots]''. Others describe the inclusion of explanatory content about the \AIA, such as ``The EU \textit{AI Act} Explained -- Blog or article section [\ldots]'', or the addition of excerpts of the regulation as source material, as in a commit that ``seeds 3 real governance excerpts (EU \textit{AI Act}, [\ldots]''. In other cases, the \AIA is mentioned in the results of a benchmark assessment, ``weighted 2.05/5, [\ldots] internally drafted rules for AI use, and explicit EU \textit{AI Act} [\ldots]''.



As for the other sources, the distribution of labels differs across repository types. Unlike in the other sources, all three repository types contain the same three labels. In \textit{Models}, \textit{Compliance and responsibility statements} account for 90\% of the commits (36), while \textit{Knowledge resources} (3; 7.50\%) and \textit{Compliance artifacts} (1; 2.50\%) are rare. In \textit{Datasets}, three of the six commits contain \textit{Compliance and responsibility statements} (50\%), two refer to \textit{Knowledge resources} (33.33\%), and one to a \textit{Compliance artifact} (16.67\%). \textit{Spaces} show the most balanced distribution: \textit{Compliance artifacts} are the most frequent label (33; 38.37\%), followed by \textit{Knowledge resources} (29; 33.72\%) and \textit{Compliance and responsibility statements} (24; 27.91\%).

\begin{rqanswer}{Answer to RQ3}
Commits mentioning the \AIA mostly belong to \textit{Spaces}, and describe three kinds of changes: adding or updating \textit{Compliance and responsibility statements}; building and documenting \textit{Compliance artifacts}, including disclaimers about limitations; and adding or modifying explanatory material or excerpts of the \AIA \textit{Knowledge resources}. No commit contains a \textit{Non-applicability claim} or a \textit{Reported alleged violation}. 
\end{rqanswer}

\section{Discussion}
\label{sec:disc}
In this section, we discuss the main observations that emerge across repository cards, discussions, and commits, and their implications for practitioners, platforms, regulators, and researchers.

\subsection{Explicit engagement with the \AIA} 

The first and most general observation concerns how rarely the \AIA is explicitly mentioned at all. Compared with the total number of repository cards, discussions, and commits on the platform, explicit mentions represent only a small fraction. Of more than 4.2 million repository cards, only 1,394 (0.033\%) contain a relevant explicit mention. The share is even smaller for community and development activity: about 0.004\% for discussions and about 0.0002\% for commits.

This is notable given the timing of our data collection. The \AIA was already in force, and the obligations for \GP models had applied for about a year. However, the regulation is still relatively new: its enforcement mechanisms, including fines, have only recently started to apply, so the community is still adapting. Moreover, compliance is a legal and therefore sensitive topic, which may often be handled outside the platform. This does not automatically imply that there is no AI Act compliance, but we note that this is a stark contrast to reporting with respect to, \eg the GDPR, where compliance reporting is common. Whether the underlying reason is that this is no primary concern within current documentation, lack of awareness for the need to document compliance, or lack of templates on how to document this, is unclear and requires further research.

\subsection{Non-applicability claims}

At the other end of the spectrum, a few repository cards argue that the \AIA does not apply to them. The reasons given vary: some owners state that no authority can certify their conformity, while others argue, based on the definitions in the Act, that certain obligations do not cover their artifact. In one case, the owner treated an indicative compute criterion from the Commission's non-binding guidelines~\cite{gpai-guidelines} as a threshold of the Act itself, and concluded that the model is not a \GP model; yet falling outside this definition only exempts a model from the \GP obligations, not from obligations that may arise from its use, \eg in high-risk or transparency-relevant contexts. These claims show that there is still uncertainty about whether and how the regulation applies, and that practitioners currently interpret it themselves. This further indicates the need for systematic, platform-wide support to check for and document AI Act compliance and possibly also better clarification by regulators regarding the definition of the applicable scope of the \AIA.

\subsection{Possible reliance on open-source exceptions} 

We believe that part of the reason for a lack of \AIA compliance reporting may be due to publishers relying on the open-source exception for models without systemic risk. However, this is not the case purely by publishing on Hugging Face. To fall under this exception, the models must also allow usage without restrictions, the modification of the weights, and publish information about the model's training and architecture. Since Liang~\etal~\cite{liang2024systematic} found that fewer than half of the models actually have a card and Longpre~\etal~\cite{longpre2024large} found high rates of license omissions in datasets, such a belief might be misguided in many cases. 

\subsection{Compliance expressed through statements} 
When the \AIA is mentioned, it is most often in the form of \textit{Compliance and responsibility statements}. In the statements we observed, owners often claim compliance, shift responsibility to the users of the artifact, or classify its risk under the regulation, frequently in a couple of sentences. This suggests that these statements focus more on who is responsible for compliance than on how compliance is achieved. Notably, the lack of information on how compliance is achieved also means that it is hard to determine if compliance is actually achieved. Reporting should be improved to at least specifically reference other parts of documentation (e.g., data documentation sections) that fulfill the compliance requirements. 

\subsection{Compliance ecosystem} 
Beyond statements about their own artifacts, a number of repositories provide resources and tools intended to help others with the \AIA. These include tools such as risk classifiers, compliance checkers, and models that audit documentation, as well as knowledge resources, such as models fine-tuned on the Act to answer questions about it, or datasets that compile relevant regulations. Many of these tools and resources are present in \textit{Spaces} and \textit{Datasets}, which may suggest that practitioners treat the \AIA not only as a constraint but also as a subject for development and for better understanding the regulation.

\subsection{The \HF community as a monitoring mechanism} 
While repository cards reflect how owners present their own artifacts, discussions show how other users react to them. \textit{Reported alleged violations} are almost absent in cards but are one of the most frequent labels in discussions, suggesting that discussions offer users a place to point out potential problems in other repositories, acting as a community monitoring mechanism. However, our data also show the limits of this form of monitoring: ten of the 15 discussions reporting alleged violations were opened by the same user, who was reported as spam by other users in three cases, and two of these discussions were hidden by the repository owners. Without this user, only five such discussions remain, which indicates that monitoring by the community is limited, not always welcomed, and can be misused.

\section{Implications}
\label{sec:implications}

Since non-compliance with the AI Act exposes publishers of resources on Hugging Face to future legal repercussions, we believe that the current large gap between published resources and actively documented compliance requires immediate action. While community awareness, guidelines to check compliance, and templates for documentation are good starting points, they still rely on proactive actions by actors publishing on Hugging Face. We believe that direct support by Hugging Face -- and other platforms where AI resources are published -- is required to effectively address this problem. 

A possible solution can be tools similar to the security check that GitHub implements by checking dependencies: an automated check could assess whether AI Act compliance reporting is required for a repository and whether reporting is already in place and inform users about the finding, prompting them to update their documentation. Such checks, combined with platform-provided templates and possibly even LLM-based support to modify existing model cards to add the information required for compliance (assuming, of course, that the underlying resources are compliant), would likely be an efficient remedy for the current lack of AI Act compliance documentation. 

Since platforms like Hugging Face, as the host of the models, datasets, and spaces might also be legally exposed through the non-compliance of their users, we believe this course of action is in the best interest of both users and platforms. The alternative would be community efforts to develop such tools and run them at scale to inform the users.

\section{Threats to Validity}
\label{sec:limitations}
In this section, we describe the threats to validity according to Wohlin~\etal~\cite{wohlin2012experimentation} that we have identified and how we handled them. 

\subsection{Construct validity}
Our analysis relies on explicit mentions of the \AIA, identified through a set of search terms (Table~\ref{tab:term-matching2}). Repositories that address requirements of the \AIA without naming it, or that refer to it with terms not included in our search, are not captured. To reduce this threat, our search was case-insensitive and treated punctuation as a word separator, so that variants such as ``ai-act'' were also matched. In addition, we included the regulation number and the less common acronym ``AIA''. Moreover, an explicit mention does not
demonstrate compliance, and the absence of a mention does not imply non-compliance. Our results therefore describe how the \AIA is referenced, not whether repositories comply with it.

\subsection{Internal validity}
The labeling of items involves subjective judgment. To mitigate this, two authors labeled the records independently, and disagreements were resolved through discussion. The lowest agreement across the three sources was Cohen's $\kappa = 0.79$, which is considered substantial according to Landis and Koch~\cite{landis1977measurement}, or moderate according to the more restrictive interpretation by McHugh~\cite{mchugh2012interrater}. Furthermore, the examples reported in Section~\ref{sec:results} are illustrative and do not constitute a systematic sub-classification within each label.

Automated accounts, single users, or organizations publishing many similar repositories can produce repeated or templated content, as illustrated by the single user who opened ten of the 15 discussions reporting alleged violations. Such cases may inflate the counts for individual labels and repository types. We kept these items in our counts, since they constitute genuine references to the \AIA, and report the most prominent case explicitly in our results. To assess the extent of repetition within repositories, we compared the number of labeled items with the number of unique repositories involved, which differed only slightly. However, we did not systematically check whether items stem from automated accounts or from the same owners or organizations, so repeated content across
repositories cannot be ruled out.

Finally, our analysis is restricted to repositories that are public at the time of collection, i.e., our results do not cover private, gated, and deleted repositories, as well as discussions hidden by the owners. Especially gated repositories are an interesting case that should be addressed by future work, as one reason for the gating might be restricting access to achieve \AIA compliance. However, since gated repositories require complying with terms of service specific to each gated repository, this analysis is out of our scope and we cannot generalize our findings to these cases.

\subsection{External validity}
Our study is limited to \HF. Although it is the largest platform for sharing models, datasets, and AI applications, practices on other platforms may differ. In addition, our data reflect a single snapshot up to 3~September~2026; we did not analyze how references to the \AIA evolve over time. Given that the regulation, its guidelines, and the obligations that apply are still evolving, the results may change as further obligations enter into application. Finally, we only considered English content, which excludes references in other EU languages, such as German, Spanish and Italian.

\section{Conclusions and Future Work}
\label{sec:conclusions}
Within this paper, we presented a first look at how the \HF community explicitly references the \AIA, about one month after the transparency obligations of Art.~50 started to apply and fines for \GP providers became possible. We found that explicit references are rare: out of more than 4.2 million repository cards, one million discussions, and 73 million commits, only 1,394 cards, 41 discussions, and 132 commits actually refer to the Act. When the Act is mentioned, this is mostly through short \textit{compliance and responsibility statements} (1,128 of the 1,394 cards), which state who is responsible for compliance rather than how compliance is achieved. Moreover, a few owners argue that the Act does not apply to them, partially based on an incorrect understanding of the regulation, and most reports of alleged violations in discussions stem from a single user. We believe that this shows that the current state of compliance documentation on \HF is not sufficient. Platforms should provide direct support, \eg through automated checks and documentation templates.

Since our study is only a first look, there are several directions for future work. Future studies should replicate our work to observe how the community's approach to AI Act compliance evolves over time. Since artifact publishers may document compliance without naming the \AIA, \eg by publishing a training-data summary based on the template of the EU AI Office or a copyright policy, deeper analysis of model cards at scale is required to assess the compliance gap accurately. Such work should also confirm whether compliance statements are indeed valid or if they are just statements without backing. Finally, tools to support compliance documentation and to check for compliance should be developed in a manner that they can be applied at scale, and their effectiveness should be evaluated.

\vspace{0.3em}
\noindent
\textbf{Data Availability:}
\ifanonymous
\else
Code and data are available at: \url{https://tinyurl.com/AIACTHF}
\fi

\vspace{0.3em}
\noindent
\textbf{Acknowledgments:}
We used Claude Code to check the manuscript for consistency. The authors reviewed and edited all generated content and take full responsibility for the content of this paper.

\bibliographystyle{IEEEtran}
\bibliography{references}

\end{document}